\documentclass[10pt, a4paper]{article}

\usepackage[]{lrec-coling2024} 
\usepackage{amsmath}
\usepackage{natbib}
\usepackage{times}
\usepackage{amssymb}
\usepackage{graphicx}
\usepackage{caption}
\usepackage{booktabs}
\usepackage{hyperref}
\usepackage{tabularx}
\usepackage{adjustbox}
\usepackage{array}
\usepackage{multirow}
\usepackage{multicol}
\usepackage{subfigure}
\newtheorem{problem}{Problem}

\title{Enhancing Court View Generation with Knowledge Injection and Guidance}

\name{Ang Li, Yiquan Wu, Yifei Liu, Fei Wu, Ming Cai\textsuperscript{\dag} \thanks{\textsuperscript{\dag} Corresponding author}, Kun Kuang\textsuperscript{\dag}} 

\address{College of Computer Science and Technology, Zhejiang University, Hangzhou, China \\
         \{liangrex, wuyiquan, liuyifei, wufei, cm, kunkuang\}@zju.edu.cn\\}

\abstract{
Court View Generation (CVG) is a challenging task in the field of Legal Artificial Intelligence (LegalAI), which aims to generate court views based on the plaintiff claims and the fact descriptions. While Pretrained Language Models (PLMs) have showcased their prowess in natural language generation, their application to the complex, knowledge-intensive domain of CVG often reveals inherent limitations. In this paper, we present a novel approach, named Knowledge Injection and Guidance (KIG), designed to bolster CVG using PLMs. To efficiently incorporate domain knowledge during the training stage, we introduce a knowledge-injected prompt encoder for prompt tuning, thereby reducing computational overhead. Moreover, to further enhance the model’s ability to utilize domain knowledge, we employ a generating navigator, which dynamically guides the text generation process in the inference stage without altering the model’s architecture, making it readily transferable. Comprehensive experiments on real-world data demonstrate the effectiveness of our approach compared to several established baselines, especially in the responsivity of claims, where it outperforms the best baseline by 11.87\%.
\\ \newline \Keywords{Natural Language Generation, Knowledge Discovery, Court View Generation} }
\begin{document}

\maketitleabstract

\section{Introduction}

In recent years, artificial intelligence has been applied in the legal domain. Legal artificial intelligence (LegalAI) focuses on applying methods of artificial intelligence to benefit legal tasks \cite{zhong-etal-2018-legal,zhong2019jecqa,yue2021circumstances,feng-etal-2022-legal, LiuWZS0WK23,wu-etal-2023-precedent}. Courts' views can be regarded as interpretations of case judgments, providing specific legal rules applicable to individual cases. Courts' views help ensure fairness and justice in judgments and Court View Generation (CVG) is considered one of the most critical functions in LegalAI.

\begin{figure}[t]
    \centering
    \includegraphics[width=0.5\textwidth]{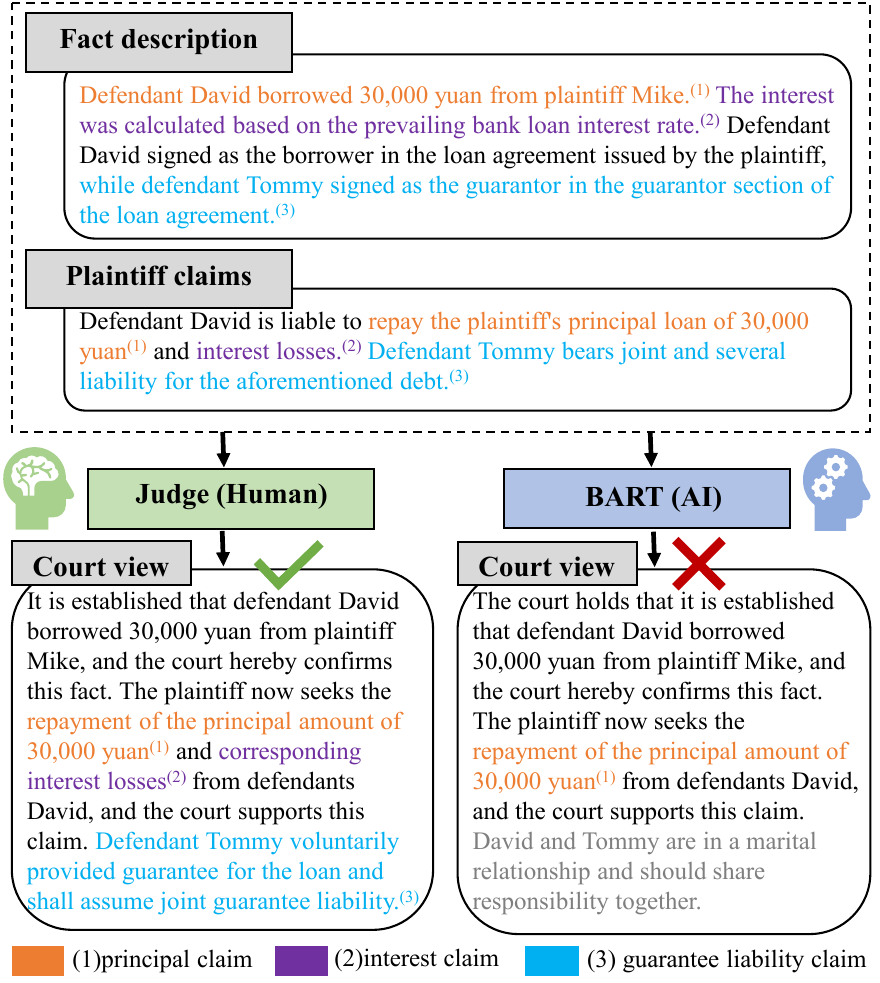}
    \vspace{-18pt}
    \caption{A real case in CVG. BART does not respond to the plaintiff's interest claim and mistakenly responds to the guarantee liability claim.}
    \vspace{-15pt}
    \label{fig:example}
\end{figure}

CVG is a distinctive natural language generation (NLG) task, which aims to generate court views based on the plaintiff claims and the fact descriptions. Existing NLG methods rely on the extensive parameters of pretrained language models (PLMs) to achieve impressive generative capabilities. These methods can be employed to address the CVG task, but their effectiveness may be compromised by insufficient domain-specific knowledge. For example, Fig. \ref{fig:example} shows the court views summarized by the judge and generated by BART \cite{lewis2019bart} in a real case. The judge provides a court view that accurately responds to the plaintiff claim based on their claim-related knowledge. However, BART might mistakenly interpret ``interest" as ``curiosity" due to its lack of understanding of claim-related knowledge, thereby failing to recognize the word's importance, resulting in the generated court view lacking the response to the interest claim. Similarly, BART also provides an incorrect response to the guarantee liability claim. In conclusion, general NLG methods may produce incomplete or erroneous court views, due to a lack of emphasis on domain-specific knowledge.

In this paper, we focus on the knowledge problem in the CVG task. To address this problem, we are still facing the following challenges: (1) Fine-tuning PLMs on domain-specific data consumes significant time and computational resources. How to inject the domain knowledge into the model efficiently? (2) There remains a gap in the model's acquisition and application of domain knowledge. How to guide the model in utilizing knowledge to fulfill the task's requirements?

To tackle the aforementioned challenges, based on the pretrained language model (PLM), we propose a Knowledge Injection and Guidance (KIG) method. To inject the domain knowledge efficiently, we propose a knowledge-injected prompt encoder, which focuses on training parameters within the prompt encoder itself. We inject two types of claim-related knowledge: keyword knowledge and label definition knowledge. Specifically, in the training stage, the knowledge-injected prompt encoder leverages keyword knowledge for its initialization. Additionally, it employs label definition knowledge to identify and emphasize claim-related information within the context. To guide the model in utilizing knowledge, we propose a generation navigator in the inference stage, which guides the PLM in generating text that incorporates useful domain knowledge. The navigator dynamically adjusts the decoding distribution, providing valuable guidance to the PLM in crafting appropriate responses to plaintiffs' claims. Importantly, it is worth noting that the navigator does not alter the architecture of the PLM and only is used in the inference stage, making it readily transferable to other models.


In order to evaluate the quality of generated court views more comprehensively, in addition to using similarity metrics, we design novel metrics to evaluate the model's ability to respond to claims. We conduct extensive experiments on a real-world dataset and demonstrate the effectiveness of our method through comparisons with multiple baselines (including ChatGPT). Our method has shown an improvement of 11.87\% in the claim response metric compared to the best-performing baseline. In summary, our work's contributions can be summarized as follows:
\begin{itemize}
\item[$\bullet$] We investigate the task of court view generation (CVG) by taking the domain knowledge into consideration. 
\item[$\bullet$] We propose a Knowledge Injection and Guidance (KIG) method, incorporating a prompt encoder for knowledge injection and a navigator for knowledge utilization guidance. The navigator holds transferability on other PLMs.
\item[$\bullet$] We design claim response metrics for the CVG task. Experimental results on a real-world dataset demonstrate the effectiveness of our method by both automatic evaluation and human evaluation. We make the code and dataset publicly available \footnote{\href{https://github.com/LIANG-star177/KIG}{https://github.com/LIANG-star177/KIG}} for reproducibility.
\end{itemize}

\section{Related Work}

\subsection{Prompt Learning}
Prompt learning is an approach that augments the input with prompt information to guide the model in adapting to downstream tasks. Researchers demonstrated its effectiveness using manually designed prompts that were intuitive and straightforward \cite{reynolds2021prompt,yu2022legal}. However, manual prompt construction can be time-consuming and labor-intensive, leading to the development of automatic methods \cite{promptmining, shin2020autoprompt}. These methods all use discrete prompts. In continuous prompt methods, they inserted trainable parameters into the input and constructed prompts in the vector space \citep{li2021prefix,liu2021p,liu2021gpt, Gu2021ResponseGW}. Additionally, studies have shown that using task-related token \cite{li2021prefix} or prior knowledge \cite{chen2022knowprompt} to initialize the prompt can enhance the performance of some tasks. In contrast to prior work's emphasis on performance in low-resource situations, our focus is on efficiently injecting domain knowledge to tackle domain-specific problems by prompt tuning. 

\begin{table*}[t]
  \centering
  \small
  \renewcommand{\arraystretch}{0.1}
  \setlength{\tabcolsep}{3pt}
  \begin{tabularx}{\textwidth}{>{\centering\arraybackslash}m{2.2cm}|>{\centering\arraybackslash}m{3cm}|>{\arraybackslash}m{10.3cm}l}
    \toprule
    \textbf{Claim Labels} & \textbf{Keywords} & \textbf{Descriptions}  \\
    \midrule
    Principal Claim & \{``principal'', ``debt'', ``borrower''\} & The requests for repayment of the initial borrowed or owed amount, excluding interest and additional charges. \\
    \midrule
    Interest Claim & \{``interest'', ``interest rate'', ``bank''\} & A borrower requests to pay the interest on a owed amount, calculated based on the agreed-upon interest rate in the loan contract. \\
    \midrule
    Spousal Joint Debt Claim & \{``spouse'', ``joint debt'', ``property division'', ``marriage''\} & One spouse seeks to divide shared debts within a marriage, often occurring during divorce or separation when assets and debts are being split. \\
    \midrule
    Guarantee Liability Claim & \{``guarantor'', ``guarantee'', ``guaranty contract''\} & A guarantor asks to fulfill their duties in a guarantee contract, typically because the borrower failed to meet their contract terms, leading to the guarantor paying the debt or fulfilling guaranteed responsibilities.\\
    \bottomrule
  \end{tabularx}
  \vspace{-5pt}
\caption{Keywords and Label definitions of each claim label.}
\label{table:claim_knowledge}
\vspace{-10pt}
\end{table*}

\subsection{Controlled Text Generation}
To make the generated text of PLMs satisfy certain constraints or serve specific purposes, controlled text generation (CTG) techniques have rapidly developed. Some researchers believe that achieving CTG requires retraining or modifying the model \cite{keskar2019ctrl}. Other researchers have achieved control effects by fine-tuning PLMs on specific corpora \cite{zhang2020dialogpt,ziegler2020finetuning,zhang2022discup}. Additionally, some researchers proposed that CTG can be achieved through straightforward post-processing techniques \cite{Yang_2021}, where the model's output is modified during the inference stage. In our work, we employ a generating navigator to ensure that generated text adheres to the attribute of ``accurate knowledge utilization", which is similar to CTG. Unlike existing methods, we recognize the impact of generation length and design a dynamic guiding mechanism.

\subsection{Court View Generation}
The task of court view generation (CVG) has spurred the development of different methods across various scenarios. \citet{yue2021circumstances} divided the factual description into the adjudging circumstance and the sentencing circumstance. They subsequently predicted the crime, generated rationales, and ultimately combined them to form the court view. \citet{wu2020biased} introduced an attention mechanism specifically designed for plaintiff claims. They also proposed a pair of counterfactual decoders, which categorize the claims into supported and unsupported, intending to highlight the significance of claims. While these works adapt the model structure to fit CVG, our approach achieves this by incorporating domain knowledge, making it better suited to the specific requirements of CVG.

\section{Problem Formulation}
\label{sec:task}
Here, we formalize our problem as follows:

\textbf{Fact Description} refers to a descriptive text that represents events that have been conclusively identified by the court. Here, we denote fact description $f =\left \{w_{1}^f,...,w_{l_f}^f \right \} $, where $l_f$ denotes its length.

\textbf{Plaintiff Claims} refer to the specific allegations and requests made by the plaintiff against the defendant. 
Here, we denote plaintiff claims $c =\left \{w_{1}^c,...,w_{l_c}^c \right \} $, where $l_c$ denotes its length. 

\textbf{Claim Labels} are categories into which plaintiff claims can be summarized. We use $m$ claim labels to encompass all the claims. We extract these labels from the plaintiff claims through word matching according to keywords \footnote{We ask legal experts to supply these keywords} in Tab. \ref{table:claim_knowledge}.

\textbf{Court View} consists of the corresponding response to the plaintiff claims and a comprehensive analysis of fact description.  We denote court view $v =\left \{w_{1}^v,...,w_{l_v}^v \right \} $, where $l_v$ denotes its length.

Then, our task could be denoted as
\begin{problem}[Court View Generation]
Given the fact description $f$ and the plaintiff claims $c$, the task is to generate the court view $v$.
\end{problem}

\begin{figure*}[t]
    \centering
    \includegraphics[width=1\textwidth]{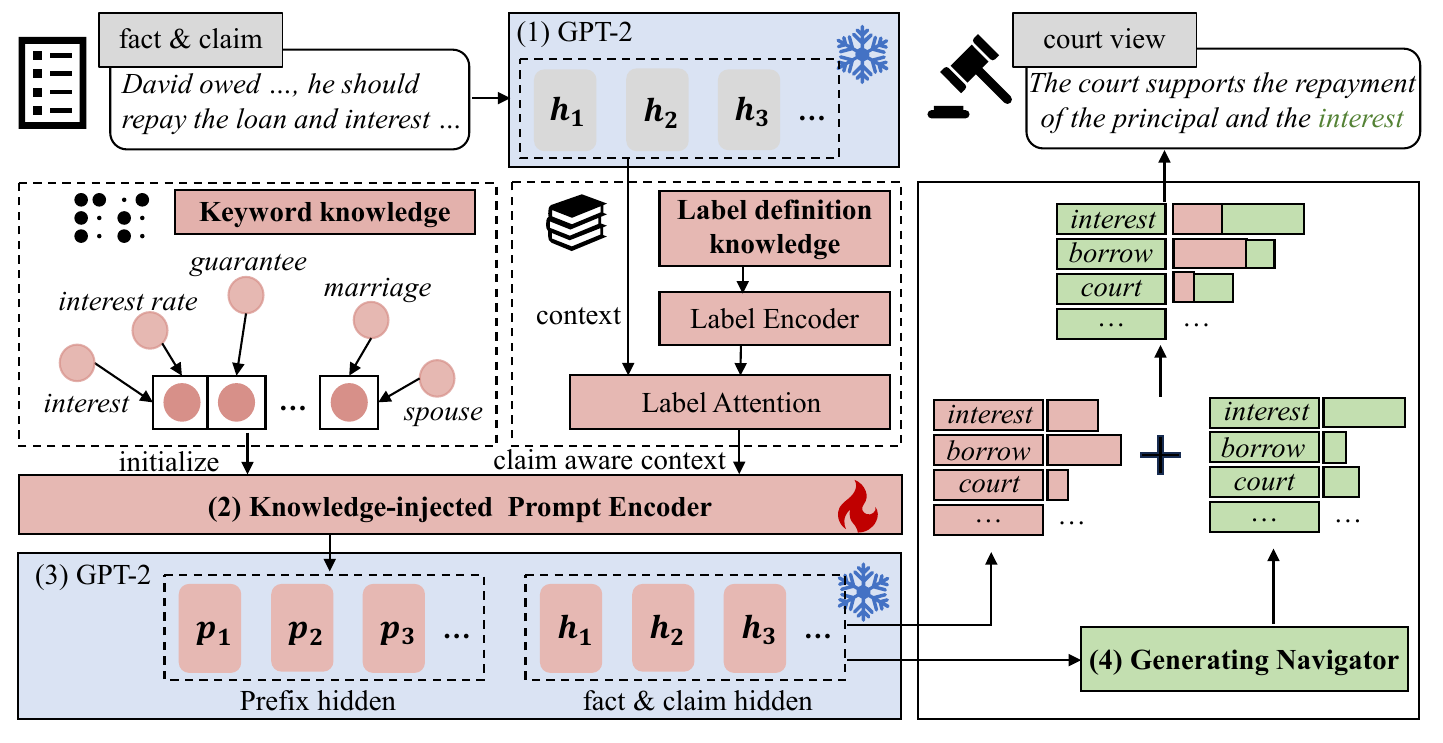}
    \vspace{-20pt}
    \caption{ The architecture of KIG. (1) is the frozen pretrained GPT-2, utilized for obtaining input context. (2) is the knowledge-injected prompt encoder, which incorporates claim knowledge to obtain claim-aware context. (3) is the same frozen GPT-2, which is used for calculating the activation of output from the knowledge-injected context. (4) is the generating navigator, guiding the model's inference process.}
    \vspace{-10pt}
    \label{fig:model}
\end{figure*}

\section{Method}
In this section, we first introduce the prompt tuning of pretrained language model (PLM), which serves as the foundational model. Then, we provide the detailed implementation of our Knowledge Injection and Guidance (KIG) method. The overall framework is shown in Fig. \ref{fig:model}.

\subsection{Preliminary}\label{Vanilla Model}
\paragraph{Prompt Tuning.} \citet{li2021prefix} train a prompt encoder that can provoke the PLM's generation ability instead of fine-tuning. Specifically, in the CVG task, a prefix $p$ of length $l_p$ is concatenated in front of the fact description, plaintiff claims, and court view denoted as $x = [p,f,c,v]$. The embeddings of the prefix are randomly initialized and denoted as $\mathcal{D}p = [e(x_1),...,e(x_{l_p})]$. Then, a trainable feedforward neural network ($MLP_\theta$) serves as a prompt encoder, transforming the prefix into the sequence of hidden states. The hidden states, excluding those from the prefix, are obtained through encoding by a pretrained autoregressive language model ($PLM_\phi$). The activation at the $i$-th time step $h_i$ can be calculated as follows:
\begin{equation}
h_i = \begin{cases}
MLP_\theta(\mathcal{D}_p)[i,:],& \text{if }i<l_p,
 \\PLM_\phi (x_i, h_{<i}),& otherwise,
\end{cases}
\end{equation}

While encoding with the PLM, $h_i = [h_i^1, ..., h_i^n]$ is computed by concatenating all activation layers. $n$ is the number of Transformer layers. Then, the probability distribution of each token in the target can be calculated as follows:
\begin{equation}
\begin{aligned}
P(x_{i+1}|h_{\leq i}) = softmax(W_\phi h_i^n)
\end{aligned}
\label{eq:distribution}
\end{equation}
where $W$ is a pretrained matrix that maps $h_i^n$ to logits over the vocabulary.

Despite the frozen parameters in the PLM, the activations introduced by the prefix naturally influence the subsequent activations of the PLM, enabling prompt tuning to achieve notable performance in low-resource scenarios. To leverage the efficiency offered by prompt tuning and address the lack of domain knowledge, we introduce a knowledge-injected prompt encoder.

\subsection{Knowledge-injected Prompt Encoder}
\label{prompt encoder section}
To efficiently address the CVG task, we incorporate two types of claim-related knowledge into the prompt encoder: keyword knowledge for initialization and label definition knowledge for label attention. The keywords and label definitions for each claim label are provided in Tab. \ref{table:claim_knowledge}.

\subsubsection{Keyword-based Initialization}
To enable the model to capture fine-grained claim information, we design a method to inject knowledge into the initialization of prompt embedding. Specifically, we first extract the representations of keywords in PLM's embedding layer and then aggregate these representations of each claim label as follows:
\begin{equation}
\hat{\mathbf{e}}_i=\sum_i^{K} \phi_i^{cl} \cdot \mathrm{E} (\mathrm{S}_i),
\end{equation}
where $K$ is the key words' number of the $i$-th claim label and $\mathrm{E}$ is the word-embedding layer of the PLM. $\mathrm{S}_i$ is the set of keywords for the $i$-th claim label, while $\phi_i^{cl}$ is the frequency distribution of these keywords. These claim label embeddings are then concatenated to initialize the embedding of the prompt encoder $\hat{\mathcal{D}_p} = [\hat{\mathbf{e}}_1,...,\hat{\mathbf{e}}_{m}]$. So the prefix length $l_p$ is equal to the number of claim labels $m$. 

\subsubsection{Label Attention}
We notice that injecting semantic knowledge of claim labels can enhance the model's ability to learn claim-related information. Since every claim label has a detailed definition in the Code of Law, we encode them using Transformer networks and a mean pooling operation to obtain the initial semantic claim label representation $C \in \mathbb{R}^{ m \times d}$. 

To allow the prompt encoder to sense different contexts, we concatenate the fact description $f$ and the plaintiff claims $c$ and inputs them into the PLM to obtain the context hidden $h_{fc} \in \mathbb{R}^{(l_f+l_c) \times d}$.  We use the context hidden to query the relevant semantics of the claim label representation. Specifically, we first calculate the similarity score between the context hidden $h_{fc}$ and the $i$-th claim label representation $C_i$ as follows:
\begin{equation}
    \alpha _i= {h_{fc}}^T W_c C_i,
\end{equation}
where $W_c$ is a trainable matrix. We calculate the weighted semantic representations of claim labels based on the corresponding similarity scores:
\begin{equation}
C' = \sum \frac{exp(\alpha_i )}{ {\textstyle \sum_{j=1}exp(\alpha_j )} } C_i
\end{equation}
We further add the input's context hidden with the most relevant semantic representation of the claim labels, obtaining the claim-aware context hidden:
\begin{equation}
h_{fc}' = h_{fc} + C'
\end{equation}

Since the MLP in prompt tuning \cite{li2021prefix} can only encode prefixes, in order to combine both claim-aware context and prefix information simultaneously, we replace the MLP with an autoregressive model $R_\theta$ as the prompt encoder. It can be expressed as follows:
\begin{equation}
h_i' = \begin{cases}
R_\theta(\hat{\mathcal{D}_p})(x_i,h_{fc}',h_{<i}),& \text{if }i<l_p,
 \\PLM_\phi (x_i, h_{<i}),& otherwise.
\end{cases}
\end{equation}

Therefore, we predict the probability distribution of each token similar to Eq. \ref{eq:distribution}.

\subsection{Generating Navigator}   
In the inference stage, we employ a generating navigator to guide the utilization of domain knowledge. Specifically, it adjusts the decoding probability distribution of the PLM to generate court views that incorporate appropriate knowledge. We will now explain the principles and the guiding process.

\vspace{-8pt}
\subsubsection{Principles}
Inspired by \citet{Yang_2021}, we explain the principles as follows:

If the generative model needs to generate a court view $v$ of length $l_v$, with the attribute $a$, it can be modeled as follows:
\begin{equation}
P(v|a)=\prod_{i=1}^{l_v} P(x_i|x_{1:i-1},a),
\end{equation}
where $x_i$ is the $i$-th token, we can utilize Bayesian decomposition to obtain:
\begin{equation}
P(x_i|x_{1:i-1},a)\propto P(x_i|x_{1:i-1})P(a|x_{1:i}),
\end{equation}
where $P(x_i|x_{1:i-1})$ is the distribution of the next token. $P(a|x_{1:i})$ can be modeled using a classifier that predicts the attribute $a$\footnote{In our task, $a$ is including the correct claim label.}.

Therefore, we can employ a claim label classifier as the generating navigator to assign the recommendation score for the next token and its training details can be found in Sec. \ref{sec:cvg-training}.

\vspace{-8pt}
\subsubsection{Recommendation Score Calculation}
Now we present the calculation process of the recommendation score. Given a generated sequence $x_{1:i-1}$, to obtain the navigator's recommendation score distribution $\phi_i^{s}$ for $i$-th token, we evaluate the normalized Jaccard similarity between the predicted label $a$ and the true label $\hat{a}$ as follows:
\begin{equation}
\phi_i^{s} = softmax(\frac{|a_i\cap \hat{a}|}{|a_i\cup  \hat{a}|} ).
\end{equation}
Moreover, we consider the impact of the generation length. When the generation length is short, the navigator receives little information, resulting in poor performance and unreliable guiding capability. We design a dynamic guidance method based on the generation length $l$, and the new scoring calculation method is as follows:
\begin{equation}
\tilde{\phi_i^{s}} = \frac{\phi_i^{s}} {1 + exp((k-l)/\mu )},
\end{equation}

where $k$ denotes the starting position of the guidance, $\mu$ controls the variation speed of the guiding strength. As the generation length $l$ increases, the guiding strength becomes stronger. 
\subsubsection{Inference with Guidance}
In the inference stage, we concatenate the fact description and the plaintiff claim together, obtaining a new input for the model to calculate the next token's distribution $\phi_i^{g}$. Simultaneously, we employ the generating navigator to calculate the recommendation score distribution $\tilde{\phi_i^{s}}$.
Then, we aggregate the two distributions to yield the desired distribution $\bar{\phi_i^{g}}$ for $i$-th token:
\begin{equation}
\tilde{\phi_i^{g}} = \phi_i^{g} +\lambda \tilde{\phi_i^{s}},
\end{equation}
where $ \lambda $ is the guiding strength coefficient. Subsequently, we obtain the court view by selecting the token with the highest probability distribution.

\subsection{Training}\label{sec:cvg-training}
\paragraph{Training for Knowledge-injected Prompt Encoder.}
We keep the PLM's parameters frozen while training the prompt encoder to make the generated texts align with the real court views. Our training objective aims to minimize the following loss function:
\begin{equation}
\mathcal{L(\theta )} = - \sum_{i=l_f+l_c+1}^{N}\log p(\phi ,\theta )(x_i|h_{<i}),
\end{equation}
where $\phi$ is the frozen parameters of the PLM and $\theta $ is the trainable parameters of the prompt encoder.

\paragraph{Training for generating navigator.}

Suppose we have a dataset $\{ (x_{1:l_v}, \hat{y}) \}$, where $x_{1:l_v}$ is a court view, and $\hat{y}$ is the corresponding claim label. Since the navigator needs to predict at each time step during the decoding process, we separate each sample $x_{1:l_v}$ into individual sub-samples $x_{1:i}$. 

We train a multi-label classifier to predict the claim labels and compare the predicted label $y$ with $\hat{y}$. We use cross-entropy loss \cite{NamKMGF14} for optimizing the loss in multi-label classification:
\begin{equation}
\mathcal{L}_N= - \sum_{i=1}^{m}\left(y_{i} \log \left(\hat{y}_{i}\right)\right) +\left(1-y_{i}\right) \log \left(1-\hat{y}_{i}\right),
\end{equation}
where predicted label for the $i$-th claim is $y_i \in [0, 1]$, the true label is $\hat{y}_i\in [0, 1]$, and $m$ is the number of claim labels. We utilize GPT-2 to train the multi-label classifier, enabling us to compute and optimize gradients for all sub-samples in a single forward-backward process. 


\section{Experiment}

\subsection{Dataset Description}
We conduct experiments on a public civil cases dataset \citep{wu2020biased}. To ensure data quality, we set the minimum length for fact description, plaintiff claim, and court view to 20, while the maximum lengths are set to 400, 200, and 400, respectively. 
We randomly split the dataset into training, validation, and testing sets in an 8:1:1 ratio. The original dataset does not contain claim labels and we extract them using word matching, as shown in the Sec. \ref{sec:task}. The details of the dataset are presented in Tab. \ref{tab:data statistics}.
\begin{table}[t]
\small
\centering
\begin{tabular}{@{}ll@{}}
\toprule
\textbf{Type}          & \textbf{Result} \\ \midrule
\# Sample              &  41693             \\
Avg. \# Tokens in Fact Description &  177.5            \\
Avg. \# Tokens in Plaintiff Claims    &  76.3              \\
Avg. \# Tokens in Court View &  205.7  \\
Avg. \# numbers of claim labels    &  2.13              \\ \bottomrule
\end{tabular}
\vspace{-5pt}
\caption{Statistics of the dataset.}
\vspace{-10pt}
\label{tab:data statistics}
\end{table}

\begin{table*}[t]
\small
    \centering
\begin{tabular}{l|cccccc|cccc}
\toprule
\multicolumn{1}{l|}{\multirow{2}{*}{Method}} & \multicolumn{6}{c|}{Similarity Metrics}                                   & \multicolumn{4}{c}{Claim Response Metrics}                            \\
\multicolumn{1}{c|}{\multirow{2}{*}{}}    & B-1           & B-2           & B-N          & R-1           & R-2           & R-L      & Mi-F           & Ma-F           & Mi-J           & Ma-J         \\ \midrule
Transformer                 & 61.01          & 52.01          & 48.97          & 64.47          & 46.45         & 56.95          & 72.02          & 65.24          & 56.28   
    & 50.86         \\
PGN                & 61.24          & 51.25          & 47.68          & 68.16          & 46.16         & 58.09          & 67.01          & 64.96          & 52.01   
    & 48.59          \\ 
AC-NLG                & 61.72          & 52.61          & 49.71          & 68.57          & 49.46         & 59.97          & 69.03         & 66.03         & 54.70   
    & 51.59         \\ \midrule
Text-Davinci-003                & 50.02          & 35.29          & 29.11         & 50.87          & 20.32        & 30.39          & 73.26          & 62.55         & 57.80  
    & 49.34         \\
GPT-3.5-Turbo               & 53.47          & 39.38          & 32.86         & 53.25         & 23.38        & 32.46         & 79.23          & 64.76         & 65.60  
    & 53.23          \\ \midrule
BART               & 64.48         & 55.72         & 51.75         & 72.12          & 51.11       & 61.89           & 76.23          & 71.47         & 61.59         & 58.54       \\
T5          & 63.33          & 53.63        & 50.33         & 69.21         & 46.39         & 61.70        & 74.01          & 72.36          & 58.74         & 59.25          \\
GPT-2         & \underline{68.65}        & \underline{61.92}         & \underline{59.05}         & \underline{75.87}        & \underline{59.06}        & \underline{68.48}         & \underline{81.09}         & \underline{75.91}         & \underline{68.19}      & 63.20          \\ 
Prefix-Tuning         & 66.28        & 58.36         & 57.12         & 73.95        & 56.52        & 65.69         & 79.28         & 75.24        & 67.59      & \underline{63.24}        \\ \midrule
KIG        & \textbf{71.04}      & \textbf{64.80}        & \textbf{62.40}        & \textbf{77.28}        & \textbf{62.25}        & \textbf{71.11}         & \textbf{90.21}         & \textbf{87.78}         & \textbf{82.16}      &\textbf{78.40}           \\ \midrule
KIG w/o V         & 70.48       & 63.58         & 61.06        & 76.47        & 60.85        & 69.99         & 88.29       &83.65         & 79.04      & 72.91        \\
KIG w/o LA        & 69.32        & 63.16         & 60.86         & 75.89        & 60.68       & 69.85        & 88.63         & 86.48        &79.58      & 76.39      \\
KIG w/o N         & 70.45       & 64.34         & 61.96         & 77.05       & 62.09       & 70.96         & 88.08         & 85.04         & 78.70      & 74.22      \\
\bottomrule
\end{tabular}
\vspace{-5pt}
\caption{Results of court view generation, the best is \textbf{bolded} and the second best is \underline{underlined}.}
\label{tab:results}
\vspace{-5pt}
\end{table*}

\begin{table*}[t]
\small
    \centering
\begin{tabular*}{0.93\textwidth}{@{\extracolsep{\fill}}l|cc|cc}
\toprule
\multicolumn{1}{l|}{\multirow{2}{*}{Method}} & \multicolumn{2}{c|}{Similarity Metrics}                                   & \multicolumn{2}{c}{Claim Response Metrics}                            \\
\multicolumn{1}{c|}{\multirow{2}{*}{}}               & B-N                     & R-L      & Mi-F           & Ma-F      \\ \midrule
Transformer w/ N                 & 48.58(-0.39)               & 56.63(-0.32)          & 73.29(+1.27)          & 66.54(+1.30)       \\
PGN w/ N                   & 46.95(-0.73)           & 57.62(-0.47)          & 68.49(+1.38)          & 65.53(+0.57)           \\ 
AC-NLG w/ N               & 49.51(-0.20)            & 59.57(-0.40)          & 70.24(+1.21)         & 67.42(+1.39)          \\ \midrule
BART  w/ N         & 51.98(+0.23)        & 62.99(+1.10)          & 78.42(+2.19)          & 72.54(+1.07)   \\
T5  w/ N             & 51.10(+0.77)              & 62.37(+0.67)        & 75.80(+1.79)     & 73.95(+1.59)      \\
GPT-2 w/ N  &60.31(+1.26)  &69.26(+0.78) &83.55(+2.46) &79.02(+3.11) \\
Prefix-Tuning w/ N             & 58.66(+1.54)               & 67.33(+1.64)         & 81.44(+1.76)         & 77.05(+1.81)      \\ \bottomrule
\end{tabular*}
\vspace{-5pt}
\caption{Results of the baselines with the generating navigator.}
\vspace{-5pt}
\label{tab:results of CVG navigator on Baseline Model}
\end{table*}

\begin{table}[t]
\small
 \begin{center}
 \begin{tabular}{p{\dimexpr 0.4\linewidth-2\tabcolsep}|p{\dimexpr 0.3\linewidth-2\tabcolsep}<{\centering}|p{\dimexpr 0.3\linewidth-2\tabcolsep}<{\centering}}
\toprule
Method     & Flu.           & Fit.       \\ \midrule
AC-NLG     & 3.18    & 2.52    \\
GPT-3.5-Turbo &3.86 &3.82 \\
GPT-2 &4.49 &3.87 \\
KIG   & \textbf{4.52}    & \textbf{4.02}   \\
\bottomrule
\end{tabular}
\end{center}
\vspace{-10pt}
\caption{Results of human evaluation.}
\vspace{-5pt}
\label{tab:results of human evaluation}
\end{table}

\begin{table}
    \centering
     \begin{tabular}{p{\dimexpr 0.35\linewidth-2\tabcolsep}|p{\dimexpr 0.2\linewidth-2\tabcolsep}<{\centering}p{\dimexpr 0.2\linewidth-2\tabcolsep}<{\centering}p{\dimexpr 0.25\linewidth-2\tabcolsep}<{\centering}}
    \toprule
         \multicolumn{1}{l|}{\multirow{2}{*}{Model}}&  Tuned Params&  Storage Space& Training Time\\ \midrule
         Finetuning&  118M&  464M& 1.4 hours\\
         KIG&  18M&  72M& 0.5 hours\\ \bottomrule
    \end{tabular}
    \vspace{-5pt}
    \caption{Efficiency Comparison of different methods on GPT-2. The training time is tested on 2 3090 GPUs.}
    \vspace{-5pt}
    \label{tab:my_label}
\end{table}

\subsection{Evaluation Metrics}
\subsubsection{Automatic Evaluation}
\paragraph{Similarity metrics.}
We use two commonly used metrics in NLG tasks. (1) \textbf{BLEU} \cite{papineni2002bleu}, measures the extent to which generated content matches the n-grams of the reference text. We used BLEU-1, BLEU-2 and BLEU-n, which is the average value of BLEU-1 $\sim$ 4. (2) \textbf{ROUGE}\footnote{https://pypi.org/project/rouge/}, is another evaluation metric for text generation. In particular, we utilized ROUGE-L \cite{lin2004rouge} to measure the longest common subsequence (LCS).

\paragraph{Claim response metrics.}

In the Controlled Text Generation (CTG) task, \citet{Yang_2021} trained a classifier to predict the formality of translated text. Similarly, we train a claim label classifier on the training data to evaluate whether the generated court views accurately respond to the plaintiff claims. Specifically, we input the court views generated by the model into the classifier and obtain predicted claim labels. After statistics, the Micro-F1 and Macro-F1 of the classifier are 95.32\% and 94.38\%, respectively, which illustrates the classifier can effectively compute claim response metrics. We use the F1 score (Mi-F, Ma-F) and Jaccard similarity coefficient (Mi-J and Ma-J) to measure the alignment between predicted labels and true labels, where Mi refers to the micro average and Ma refers to the macro average. 

\subsubsection{Human Evaluation}
To further evaluate the effectiveness of KIG, we conduct a human evaluation. We randomly select 200 civil cases from the test set and shuffle them to ensure fairness. We invite ten annotators with legal backgrounds to evaluate these results refer to the ground truth from two perspectives:
(1) \textbf{Fluency.} The annotators rate the fluency of court views on a scale of 1-5.
(2) \textbf{Fitness.} The annotators assess the alignment between the court view and the plaintiff claims assigning a score between 1-5.

\subsection{Baseline Methods}
First, we implement several NLG methods that are trained from scratch.
(1) \textbf{Transformer} \cite{vaswani2017attention}, is a common method that leverages self-attention mechanisms.
(2) \textbf{PGN} \cite{see2017get}, is a sequence-to-sequence model that designs a pointer mechanism to handle out-of-vocabulary words.
(3) \textbf{AC-NLG} \cite{wu2020biased} is a civil CVG method, that mitigates the bias of data.

Recently, there has been a surge in the popularity of large language models (LLMs). We incorporate a prompt (``Please generate a court view based on the following facts and the plaintiff claims") before the input and utilize their interfaces to generate court views. (4) \textbf{Text-Davinci-003}, can generate high-quality text outputs with its large-scale pretraining on diverse text data. 
(5) \textbf{GPT-3.5-Turbo}\footnote{GPT-3.5-Turbo is the model that powers ChatGPT}, is an upgraded version of the GPT-3 model, and exhibits impressive NLG capabilities. 

In addition, we implement several PLMs-based methods. (6) \textbf{GPT-2}, is fine-tuning the pretrained GPT-2 \footnote{https://huggingface.co/uer/gpt2-chinese-cluecorpussmall} model on civil court cases. (7) \textbf{BART} \cite{lewis2019bart} is a bidirectional autoregressive Transformer model that performs well in various NLP tasks. (8) \textbf{T5} \cite{raffel2020exploring} is a text generation model based on the Transformer architecture and follows a text-to-text transfer learning paradigm. (9) \textbf{Prefix-Tuning} \cite{li2021prefix} is a prompt learning method that employs a prompt encoder to encode prefixes. We use the same pretrained GPT-2 model as the base model to implement this method.

Lastly, we conduct ablation experiments on our approach. (10) \textbf{KIG w/o V} removes the knowledge prompt initialization from the prompt encoder and instead initializes it with random embeddings. (11) \textbf{KIG w/o L} removes the label attention from the prompt encoder and uses the context hidden of the input. (12) \textbf{KIG w/o N} removes the generating navigator in the inference stage, generating court views solely based on the decoder. 

\subsection{Experiment Details}
\label{app:setup}
In prompt initialization, we set the prefix length to the number of claim labels.  To maintain alignment, we set the same prefix length in the prefix-tuning method. When using the generating navigator to guide the generation process, we empirically set the range for changing the probability distribution to 10 \cite{Yang_2021}. This means that the top 10 tokens with the highest probabilities are scored by the generating navigator and re-ordered.  We explore different hyperparameters of our model, and our work reports the results obtained with the best-performing parameters, where  $\lambda$ is 6, $k$ is 50, and $\mu$ is 10. To assess the efficiency of our KIG method, we compared the training costs between fine-tuning and KIG in Tab. \ref{tab:my_label}.

\begin{figure*}[t]
    \centering
    \includegraphics[width=1.0\textwidth]{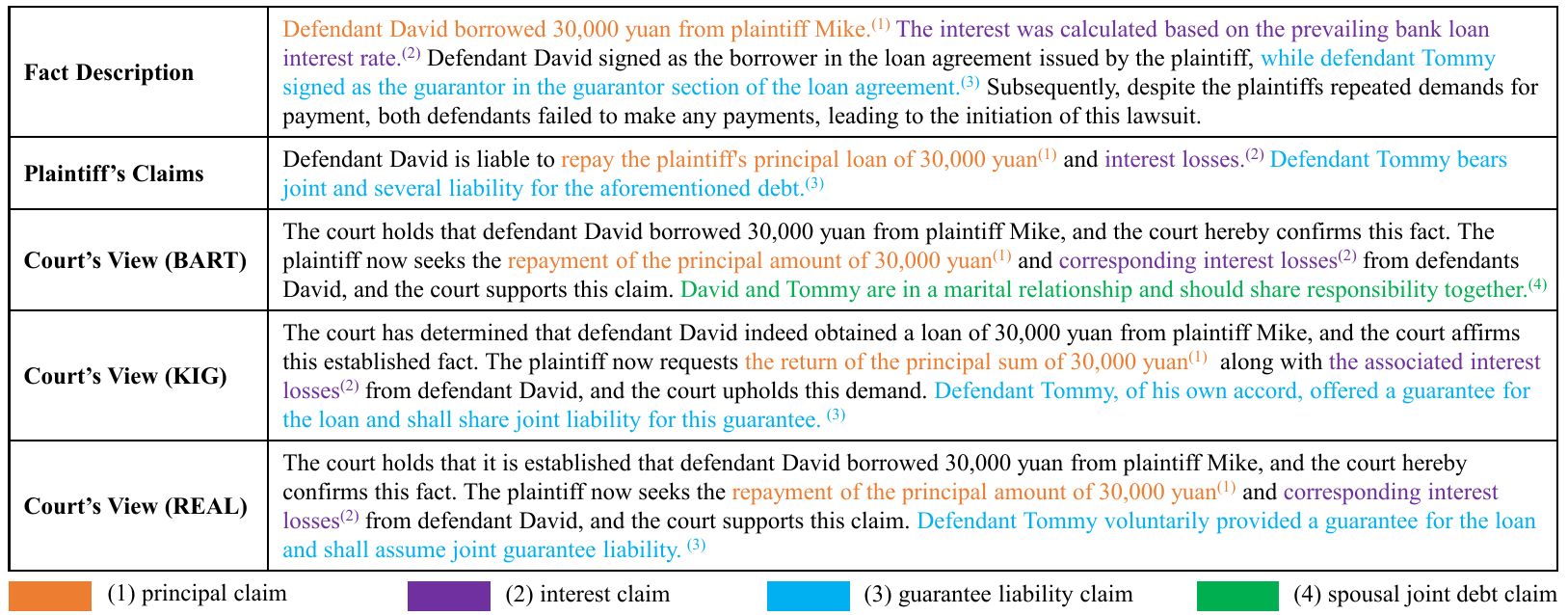}
    \vspace{-20pt}
    \caption{Case study.}
    \label{fig:case study}
    \vspace{-10pt}
\end{figure*}
\subsection{Experimental Results}
\paragraph{Results of court view generation:}
Tab. \ref{tab:results} shows results of CVG in automatic metrics. We have the following observations: (1) AC-NLG performs the best among the methods trained from scratch, but the PLMs-based methods achieve better results, with GPT-2 demonstrating the best performance among all baselines. (2) Our method outperforms the baselines across all metrics. KIG outperforms GPT-2 by 3.35\% in the B-N and by 11.87\% in the Ma-F. (3) The interface-based methods perform poorly in similarity metrics but excel in claim response metrics, which demonstrates that while LLMs may lack legal knowledge, they possess exceptional comprehension abilities. (4) Compared to GPT-2, prefix-tuning performs worse, which proves that simply replacing fine-tuning with prompt learning leads to a performance decline. (5) In the ablation experiments, KIG w/o V, KIG w/o LA, and KIG w/o N perform worse than KIG in all metrics, demonstrating the effectiveness of our approach. Specifically, knowledge prompt Initialization significantly improves the response metric, while label attention noticeably enhances the similarity metric. 

\vspace{-5pt}
\paragraph{Results of the baselines with the generating navigator:}
From Tab. \ref{tab:results of CVG navigator on Baseline Model}, we can conclude that:
(1) The generating navigator improves the claim response metrics across all baselines, indicating its versatility and transferability.
(2) In terms of the similarity metric, the generating navigator progressively improves the PLMs-based methods but slightly decreases the methods trained from scratch. We believe this is because the navigator is the PLM architecture and is better aligned with the PLM generator.
In summary, the generating navigator can be easily transferred to other models to meet the requirements of specific tasks.

\vspace{-5pt}
\paragraph{Results of human evaluation:}

From Tab. \ref{tab:results of human evaluation}, it can be seen that: (1) Due to the lack of rich language knowledge, AC-NLG performs worse than methods that involve PLMs. (2) Our method KIG achieves the best performance in fluency and fitness, which is consistent with the results in the automatic evaluation. (3) The Kappa coefficient $\kappa$ between any two annotators is greater than 0.8, indicating the consistency of the human evaluation.

From the above observations, our method can generate high-quality court views. Specifically, it can provide effective responses to the plaintiff claims while aligning with true court views.

\begin{figure}[t]
  \centering
  \begin{subfigure}
    \centering
    \includegraphics[width=1.0\linewidth]{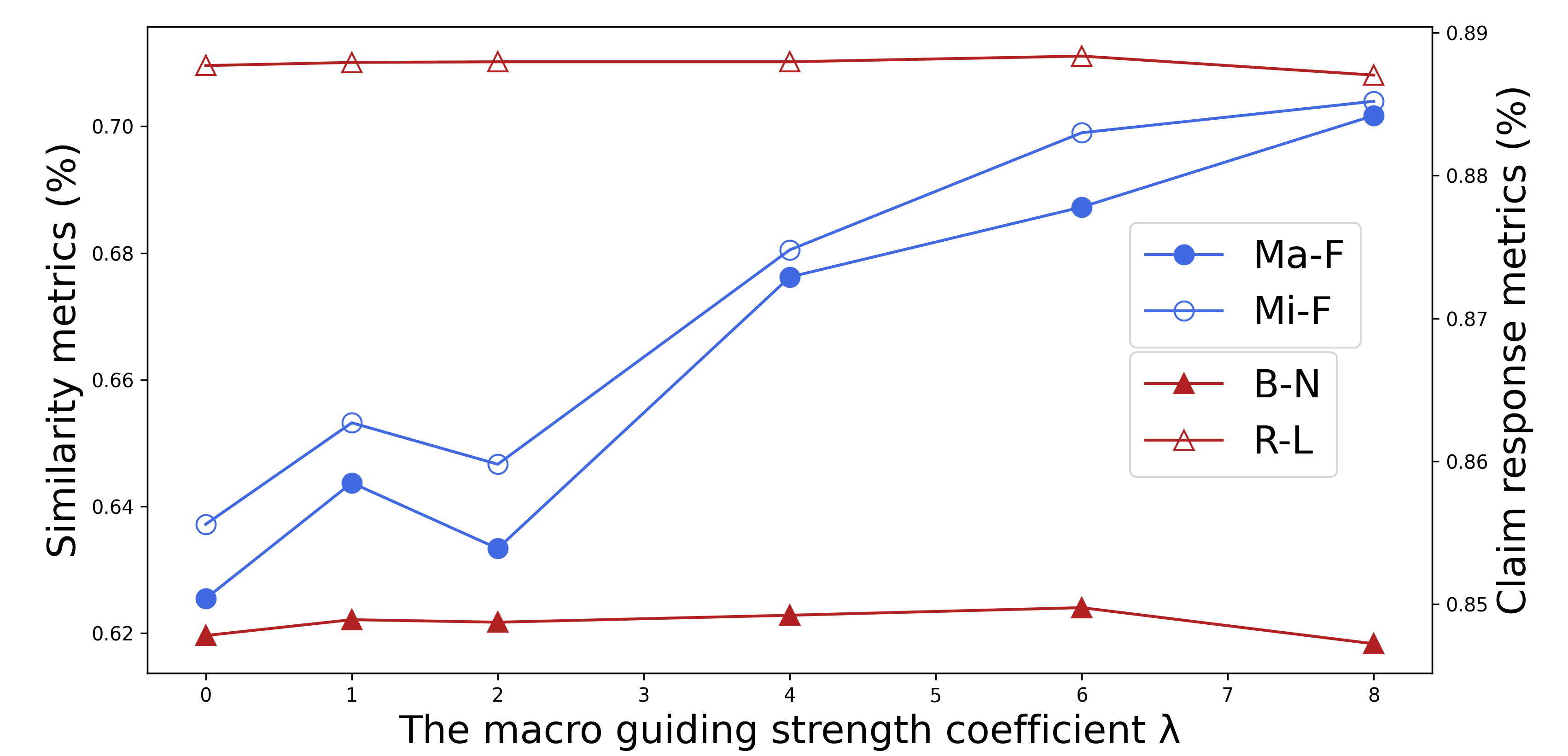}
  \end{subfigure}%
  \begin{subfigure}
    \centering
    \includegraphics[width=1.0\linewidth]{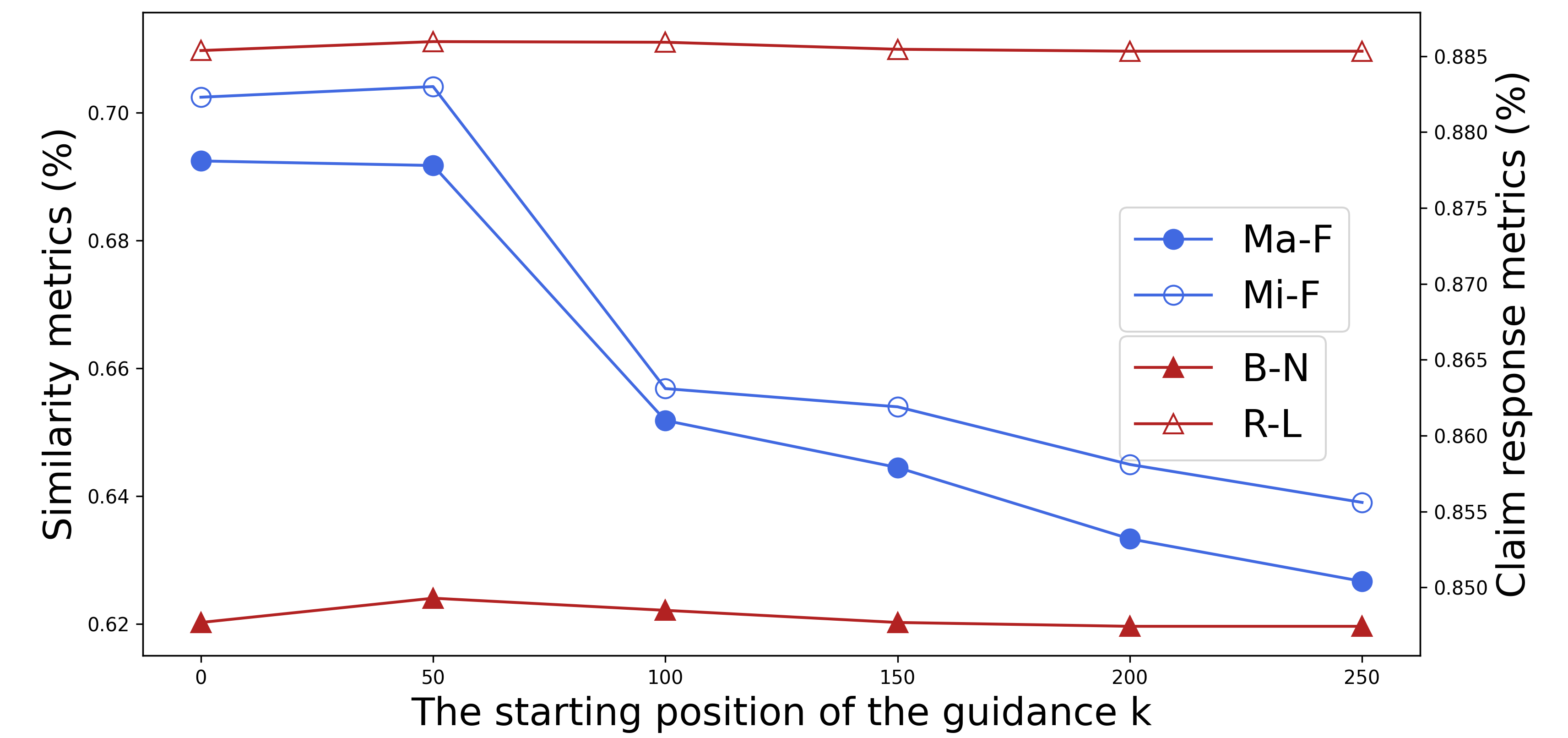}
  \end{subfigure}
  \vspace{-20pt}
  \caption{The upper shows the impact of guiding strength coefficient $\lambda$ and the lower shows the impact of  the guidance starting position $k$.}
  \vspace{-10pt}
  \label{fig:lambdak}
\end{figure}

\vspace{-5pt}
\paragraph{The impact of hyperparameters on performance:}
From the upper of Fig. \ref{fig:lambdak}, it can be observed that as the coefficient $\lambda$ increases, the claim response metrics gradually improve, but the similarity metrics start to decrease after $\lambda=6$. This indicates that excessive adjustment by the generating navigator leads to a decrease in performance. Therefore, in our work, we set $\lambda$ to 6. From the lower, we can find that when $k$ exceeds 50, there exists a noticeable decline in the guidance ability of the generating navigator. This suggests that the court view starts responding to the claims at around 50 tokens. In our work, we set $k$ to 50.

\subsection{Case Study}
Fig. \ref{fig:case study} displays court views in a real legal case, generated by the PGN, BART, and our method KIG, alongside the real court view. In this case, PGN failed to address the interest claim, while BART not only omitted a response to the interest claim but also incorrectly interpreted the guarantee liability claim as a spousal joint debt claim. In contrast, KIG accurately addressed all claims and maintained a similarity to the real court view.

\vspace{-5pt}
\section{Conclusion}
In this paper, based on the PLM, we propose an efficient \textbf{K}nowledge \textbf{I}njection and \textbf{G}uidance (KIG) method for the CVG task. Our method integrates claim-related knowledge by prompt tuning and introduces a generating navigator to guide the generation process. Through comprehensive experiments on a real-world dataset, we demonstrate the effectiveness of our method in generating court views. Our approach is based on GPT-2 but does not modify the structure of GPT-2, so our method has the potential to be generalized to other LLMs, such as ChatGLM \footnote{https://github.com/THUDM/ChatGLM-6B}, which also use prompt tuning.

\section{Ethical Discussion}
\label{app:discussion}
It is crucial to consider the ethical implications of CVG using artificial intelligence. First, Our method is constructed on a Chinese legal scene, and if it is to be applied to other countries' legal systems, adaptive adjustments need to be made. When dealing with a large amount of legal data, which involves personal privacy, it's crucial to adhere to privacy protection regulations \cite{XU202366}. Another concern is that models can inadvertently perpetuate biases from training data. Therefore, it is essential to mitigate model biases to ensure fairness.

Furthermore, while KIG can automate CVG, it is necessary to preserve the role of legal professionals in critically evaluating and interpreting the generated outputs. This work should serve as a tool to assist legal professionals rather than replace their expertise and judgment. So far, users and stakeholders should have a clear understanding of the potential limitations of this algorithm and it cannot be directly applied to real-world scenarios.


\section{Limitations}
In this section, we discuss the limitations of our work as follows:
\begin{itemize}
\item The KIG method enhances the understanding of litigation claims by injecting prior knowledge. To enhance performance, consideration could be given to appropriate feature extraction techniques \cite{community_preserve}. Moreover, law is a constantly evolving domain. Ensuring the timeliness of knowledge updates may pose challenges.

\item The KIG method introduces a generating navigator to explicitly guide the generator's focus on claims. However, this explicit guidance often requires extra training and data.
\end{itemize}

\section{Acknowledgements}
This work was supported in part by National Key Research and Development Program of China (2022YFC3340900), National Natural Science Foundation of China (No. 62376243, 62037001, U20A20387), the StarryNight Science Fund of Zhejiang University Shanghai Institute for Advanced Study (SN-ZJU-SIAS-0010) and Project by Shanghai AI Laboratory (P22KS00111).
 
\nocite{*}
\section{Bibliographical References}\label{sec:reference}

\bibliographystyle{lrec-coling2024-natbib}
\bibliography{lrec-coling2024-example}

\end{document}